\documentclass[10pt,twocolumn,letterpaper]{article}

\usepackage[pagenumbers]{cvpr}

\usepackage{bm}
\usepackage{tikz}  
\usepackage{multirow}
\usepackage{colortbl}
\usepackage{bbding}  
\usepackage{fix-cm}  
\usepackage{arydshln}  

\newcommand{\circled}[1]{\tikz[baseline=(char.base)]{\node[shape=circle,draw,inner sep=0.5pt] (char) {#1};}}  

\definecolor{americanrose}{rgb}{1.0, 0.01, 0.24}

\definecolor{cvprblue}{rgb}{0.21,0.49,0.74}
\usepackage[pagebackref,breaklinks,colorlinks,allcolors=cvprblue]{hyperref}

\title{Unleashing In-context Learning of Autoregressive Models \\ for Few-shot Image Manipulation}

\author{Bolin Lai$^{1,2\,\dagger}$ \quad
Felix Juefei-Xu$^{1}$ \quad
Miao Liu$^{1}$ \quad
Xiaoliang Dai$^{1}$ \quad
Nikhil Mehta$^{1}$ \quad
Chenguang Zhu$^{1}$ \\
Zeyi Huang$^{5}$ \quad
James M. Rehg$^{3}$ \quad
Sangmin Lee$^{4}$ \quad
Ning Zhang$^{1}$ \quad
Tong Xiao$^{1}$ \\
$^1$GenAI, Meta \quad $^2$Georgia Institute of Technology \quad $^3$University of Illinois Urbana-Champaign \\ $^4$Sungkyunkwan University \quad $^5$University of Wisconsin–Madison \\
{\tt\small bolin.lai@gatech.edu \ \{felixu,miaoliu,xiaoliangdai,nikhilmeht,chezhu,ningzhang,xiaot\}@meta.com} \\
{\tt\small zeyihuang@cs.wisc.edu \ jrehg@illinois.edu \ sangmin.lee@skku.edu} \\
}

\begin{document}

\twocolumn[{
\renewcommand\twocolumn[1][]{#1}
\maketitle
\begin{center}
\vspace{-0.7cm}
\includegraphics[width=\linewidth]{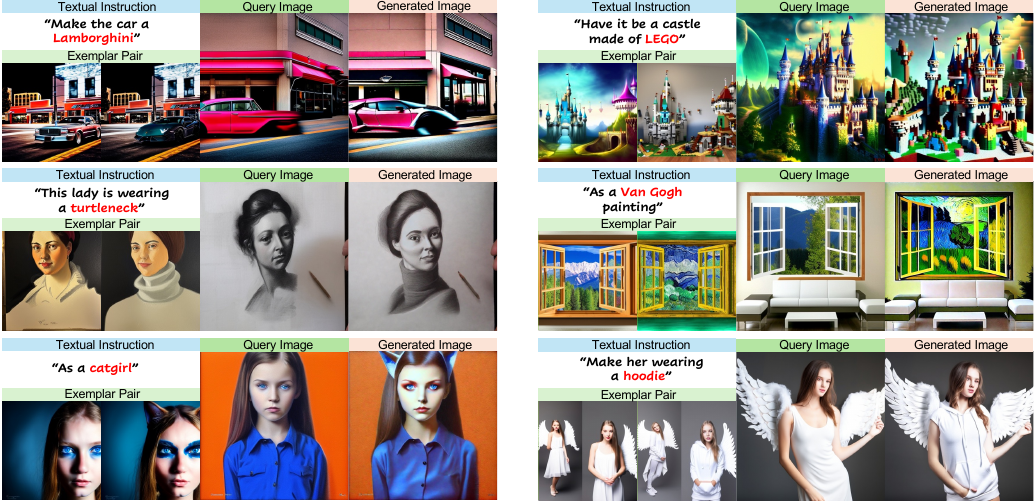}
\vspace{-0.7cm}
\captionof{figure}{When learning a new image manipulation operation that is \textbf{\textit{unseen}} in the training set (as shown above), textual instructions directly point out the subject and provide high-level semantic guidance, while exemplar images mitigate linguistic ambiguity and show more local details that are difficult to describe in language. Our proposed multi-modal autoregressive model -- \textbf{InstaManip} takes advantage of both textual and visual guidance to learn a representation of the desired transformation, and applies it to a new query image.
}
\label{fig:teaser}
\end{center}
}]

\let\thefootnote\relax\footnotetext{$^\dagger$This work was done during Bolin's internship at GenAI, Meta.}

\begin{abstract}

\vspace{-0.4cm}
Text-guided image manipulation has experienced notable advancement in recent years. In order to mitigate linguistic ambiguity, few-shot learning with visual examples has been applied for instructions that are underrepresented in the training set, or difficult to describe purely in language. However, learning from visual prompts requires strong reasoning capability, which diffusion models are struggling with. To address this issue, we introduce a novel multi-modal autoregressive model, dubbed \textbf{InstaManip}, that can \textbf{insta}ntly learn a new image \textbf{manip}ulation operation from textual and visual guidance via in-context learning, and apply it to new query images. Specifically, we propose an innovative group self-attention mechanism to break down the in-context learning process into two separate stages -- learning and applying, which simplifies the complex problem into two easier tasks. 
We also introduce a relation regularization method to further disentangle image transformation features from irrelevant contents in exemplar images. Extensive experiments suggest that our method surpasses previous few-shot image manipulation models by a notable margin ($\geq$19\% in human evaluation). We also find our model can be further boosted by increasing the number or diversity of exemplar images. Please check out our project page ({\footnotesize \href{https://bolinlai.github.io/projects/InstaManip/}{https://bolinlai.github.io/projects/InstaManip/}}).
\vspace{-0.2cm}
\end{abstract}
    
\vspace{-1cm}
\section{Introduction}
\label{sec:intro}

\begin{figure}[t]
\centering
\includegraphics[width=\linewidth]{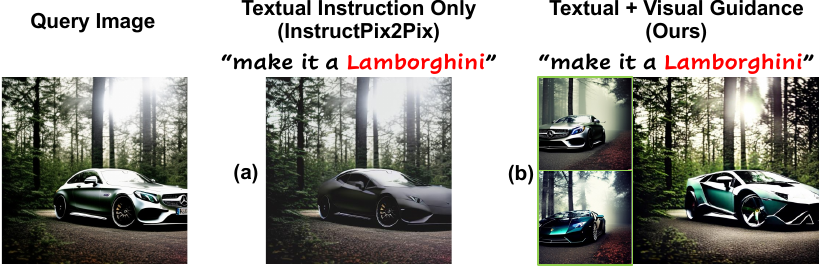}
\vspace{-0.7cm}
\caption{Comparison of InstructPix2Pix \cite{brooks2023instructpix2pix} and our model. We exclude ``Lamborghini'' from training set for both models.}
\label{fig:ip2p_vs_ours}
\vspace{-0.6cm}
\end{figure}

The recent emergence and advancement of diffusion models have greatly facilitated the boom of text-to-image generation \cite{dai2023emu,rombach2022high,ramesh2021zero,ramesh2022hierarchical,balaji2022ediff,saharia2022photorealistic,chen2024enhancing,parihar2025precisecontrol,chen2024gentron,nguyen2024swiftbrush}, which has further driven a remarkable development in text-guided image manipulation \cite{brooks2023instructpix2pix,sheynin2024emu,meng2022sdedit,hertz2023prompt,mokady2023null,zhang2023magicbrush}. However, existing models still suffer from a notable performance drop when the manipulation is difficult to articulate textually or when instructions deviate from the training data \cite{brooks2023instructpix2pix,nguyen2023visual}. For example, when we want to turn a plain car to a Lamborghini, the model may fail to correctly understand the shape and texture only from the word  ``Lamborghini'', if it is not included in training data (\cref{fig:ip2p_vs_ours}(a)). It is also hard for humans to accurately describe all details of Lamborghini in texts. Moreover, we are living in a world where new concepts constantly emerge across the Internet and social media, which are rarely covered in any training set. The generalization limitation hinders existing models from being applied in the real world. 

A straightforward solution to this problem is additionally providing a few exemplar images for the model (\ie, few-shot image manipulation as shown in \cref{fig:teaser,fig:ip2p_vs_ours}(b)), which has been studied in some recent work \cite{wang2023context,sun2023imagebrush,nguyen2023visual,zhao2024instructbrush}. All of these methods rely on the architectures of diffusion model \cite{rombach2022high} and ControlNet \cite{zhang2023adding}. However, learning from visual examples requires a strong reasoning capability to separate image-to-image transformation features from the irrelevant content in exemplar images. Diffusion models are excellent in generation, yet still weak in reasoning \cite{tang2024codi}. In contrast, autoregressive architectures, especially large language models (LLMs), have shown remarkable reasoning performance, which enables them to learn new tasks from prompts without finetuning (\ie, in-context learning) \cite{bai2024sequential,wies2023learnability,chung2024scaling,zhao2021calibrate,olsson2022context}. In this paper, we make an attempt to address few-shot image manipulation problem by harnessing the in-context learning feature of autoregressive models, specifically multi-modal large language models (MLLMs).

Prior to our work, many efforts have been made to turn an autoregressive architecture into a generalist model that can handle various visual tasks \cite{huang2024multimodal,sun2024generative,ge2024seed,sheng2024towards,bai2024sequential,lu2024unified,hernandez2024generative,fang2024puma,zhan2024anygpt,wang2024emu3}, such as visual question answering, image completion, and semantic segmentation. However, in-context learning for few-shot image manipulation with autoregressive models is still an understudied problem. In addition, few-shot image manipulation essentially consists of two stages: (1) \textit{learning} the desired transformation from textual guidance and visual examples, and then (2) \textit{applying} learned knowledge to a new query image (which has also been shown in human's learning process \cite{tenenbaum2011grow,sweller1988cognitive}). Most existing autoregressive models combine the two stages in a single step while applying in-context learning, and fully rely on self-attention to automatically model the dependence across given examples, query images and desired output. These straightforward approaches increase the problem complexity, which leads to a bottleneck in learning the desired manipulation rules and transferring to other images.

To address these issues in few-shot image manipulation, we introduce \textit{\textbf{InstaManip}}, an innovative multi-modal autoregressive architecture that models the two stages separately. Specifically, we propose a novel \textit{group self-attention} mechanism, which disentangles the learning and applying stages by splitting the input prompt into two groups and conducting self-attention in each group separately, exactly aligned with aforementioned human's cognition.
Furthermore, we introduce a \textit{relation regularization} scheme to encourage instances with similar manipulation to be encoded close to each other, which drives the model to distinguish the desired manipulation features from irrelevant image contents. The experiments show that the proposed method achieves new state-of-the-art performance when applied to \textit{unseen} image manipulation instructions. Overall, our contributions can be summarized as follows:
\begin{itemize}
    \item We introduce InstaManip, a novel autoregressive model that unleashes in-context learning capability of MLLMs for few-shot image manipulation.
    \item We propose the innovative group self-attention method that breaks down in-context learning into two stages -- learning and applying, following human's learning process. We also propose a relation regularization strategy to further separate underlying transformation rules from undesired visual features.
    \item Extensive experiments suggest that our proposed method prominently improves the in-context learning capability and outperforms existing few-shot image manipulation models. Our model is further improved by using more examples or increasing the diversity of visual prompts. 
\end{itemize}

\vspace{-0.05cm}
\section{Related Work}
\vspace{-0.05cm}

\textbf{Few-shot Image Manipulation.} Text-guided image manipulation has been widely studied since the emergence of diffusion models \cite{brooks2023instructpix2pix,sheynin2024emu,meng2022sdedit,hertz2023prompt,mokady2023null,pan2023effective,wallace2023edict,couairon2023diffedit,orgad2023editing,mirzaei2024watch,epstein2023diffusion,lin2024text,brack2024ledits++,li2023blip,lai2024lego}. Recently, few-shot learning is adopted to this problem for a better performance by using one or more exemplar image pairs as reference \cite{liao2017visual,vsubrtova2023diffusion}. Sun \etal \cite{sun2023imagebrush} propose ImageBrush, which frames query image and a pair of exemplar images into a 2$\times$2 grid, and then models their relation by a diffusion model. Wang \etal \cite{wang2023context} encode exemplar images and the query image by convolutional layers and inject the embeddings into a diffusion model through ControlNet \cite{zhang2023adding}. Their method shows excellent performance in layout-based inpainting tasks. Nguyen \etal \cite{nguyen2023visual} freeze a pre-trained InstructPix2Pix \cite{brooks2023instructpix2pix} model and finetune the condition tokens to learn the editing representations from exemplar images in CLIP space. The learned tokens can be used as conditions to edit input images. Likewise, Zhao \etal \cite{zhao2024instructbrush} propose to directly learn the representations of keys and values in each cross-attention layer. All previous work relies on diffusion models as in-context learners, which are strong in generation, yet weak in reasoning \cite{tang2024codi}. In contrast, we propose an innovative autoregressive model that can leverage the strong reasoning capability of MLLMs for few-shot image manipulation. 

\vspace{0.1cm}
\noindent\textbf{Visual In-context Learning.} In recent years, the strong in-context learning capability has been observed in LLMs \cite{brown2020language}, and subsequently extends to vision-related tasks, such as segmentation \cite{wang2023seggpt,li2024visual,wang2024explore,suo2024rethinking}, scene understanding \cite{balavzevic2023towards}, 3D point cloud modeling \cite{fang2023explore} and generalist vision models \cite{liu2024glid,huang2024multimodal,zhang2023makes,sheng2024towards,xu2024towards,xu2024improv,kim2024chameleon}. Bar \etal \cite{bar2022visual} first propose visual in-context learning by enabling models to learn from visual prompts via inpainting. Similarly, Wang \etal \cite{wang2023images} introduce Painter, a model that learns the dependence of image patches through masked image modeling, and shows strong in-context learning capability in many dense visual prediction tasks (\eg, segmentation, depth, denoising, \etc). In addition to dense prediction, the latest work shows that visual in-context learning also applies to generative models \cite{xu2024prompt,jones2024customizing,gu2024analogist,xiao2024omnigen,lu2024unified,bai2024sequential}. Sun \etal \cite{sun2024generative} develop Emu2, a unified autoregressive model showing strong in-context learning performance in text-to-image generation. Tang \etal \cite{tang2024codi} propose Codi-2, which leverages an LLM to reason from in-context examples, and uses a diffusion model to synthesize the image or audio conditioned on LLM output. Prior work mostly aims at establishing a versatile in-context learner for a variety of vision tasks, in which few-shot image manipulation is still understudied. In this work, we propose a novel method to unleash the in-context learning capability of autoregressive models on this specific problem.

\vspace{0.1cm}
\noindent\textbf{Autoregressive Models for Image Generation.} Recent studies show an increasing interests in extending LLMs into unified autoregressive models that can take in and generate image tokens directly \cite{koh2023generating,sun2024autoregressive,xiao2024omnigen,hernandez2024generative,zhan2024anygpt,dong2024dreamllm,xie2024show,team2024chameleon,fan2024fluid,tian2024visual,ma2024star,liu2024lumina,fang2024puma}. Sun \etal \cite{sun2023emu} introduce Emu, which takes in text-image interleaved prompts and synthesizes texts and images in a unified autoregressive manner. They further extend the work to Emu2 \cite{sun2024generative} and Emu3 \cite{wang2024emu3} by using discrete image embeddings and scaling up the training data. Likewise, Ge \etal \cite{ge2024making} present SEED, an LLM-based architecture that generates language and images following instructions. They also propose SEED-X \cite{ge2024seed} which is a versatile model for many vision-language tasks, such as visual question answering, open-vocabulary object localization and image editing. Zhou \etal \cite{zhou2024transfusion} replace causal masks with block masks on image tokens for a holistic understanding of images. Li \etal \cite{li2024autoregressive} find vector quantization is unnecessary in autoregressive image generation. They propose a diffusion loss to model per-token probability, achieving strong image generation performance. Most of existing work directly uses the autoregressive architectures of off-the-shelf LLMs by finetuning. How to design a novel autoregressive model for a specific problem remains to be explored, which is exactly the focus of our work.

\section{Method}

\begin{figure}[t]
\centering
\includegraphics[width=\linewidth]{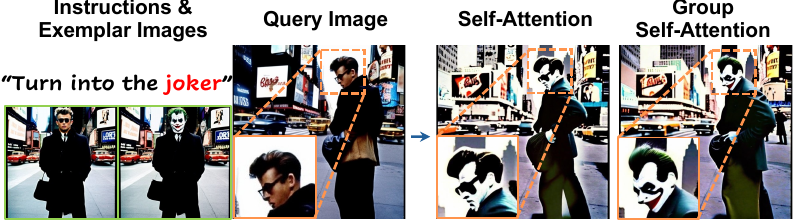}
\vspace{-0.7cm}
\caption{Comparison of the performance of plain self-attention (with causal mask) and the proposed group self-attention.}
\label{fig:gsa_vs_plain}
\vspace{-0.5cm}
\end{figure}

In few-shot image manipulation, the input is an image $\mathcal{X}$ and a textual instruction $\mathcal{T}$. We additionally use a handful of exemplar image pairs as input showing how to transform a source image $\mathcal{X}'$ to a target image $\mathcal{Y}'$. The desired output is a manipulated image $\mathcal{Y}$ following both the textual and the visual guidance. The problem is formulated as learning a distribution of $\mathcal{Y}$ conditioned on $(\mathcal{X},\mathcal{T},\mathcal{X}',\mathcal{Y}')$:
\vspace{-0.2cm}
\begin{equation}\label{eq:fewshot_formulation}
    P(\mathcal{Y}|\mathcal{X},\mathcal{T},\mathcal{X}',\mathcal{Y}').
\vspace{-0.2cm}
\end{equation}
Early studies in cognitive science \cite{tenenbaum2011grow,sweller1988cognitive} reveal that human brains follow a 2-stage learning mechanism when learning a new skill from examples -- abstracting high-level concepts from concrete examples, and then applying the learned knowledge to new cases. Inspired by this, we introduce a variable $\mathcal{Z}$ to denote abstract manipulation features, which is independent from the query image $\mathcal{X}$. Then the problem formulated in \cref{eq:fewshot_formulation} is broken down into two stages:
\vspace{-0.2cm}
\begin{equation}\label{eq:2stage_formulation}
    P(\mathcal{Y}|\mathcal{X},\mathcal{T},\mathcal{X}',\mathcal{Y}') = \underbrace{P(\mathcal{Z}|\mathcal{T},\mathcal{X}',\mathcal{Y}')}_{\circled{1}} \cdot \underbrace{P(\mathcal{Y}|\mathcal{X},\mathcal{Z})}_{\circled{2}},
    \vspace{-0.2cm}
\end{equation}
where \circled{1} corresponds to the learning stage and \circled{2} indicates the applying stage.

\begin{figure*}[t]
\centering
\includegraphics[width=\linewidth]{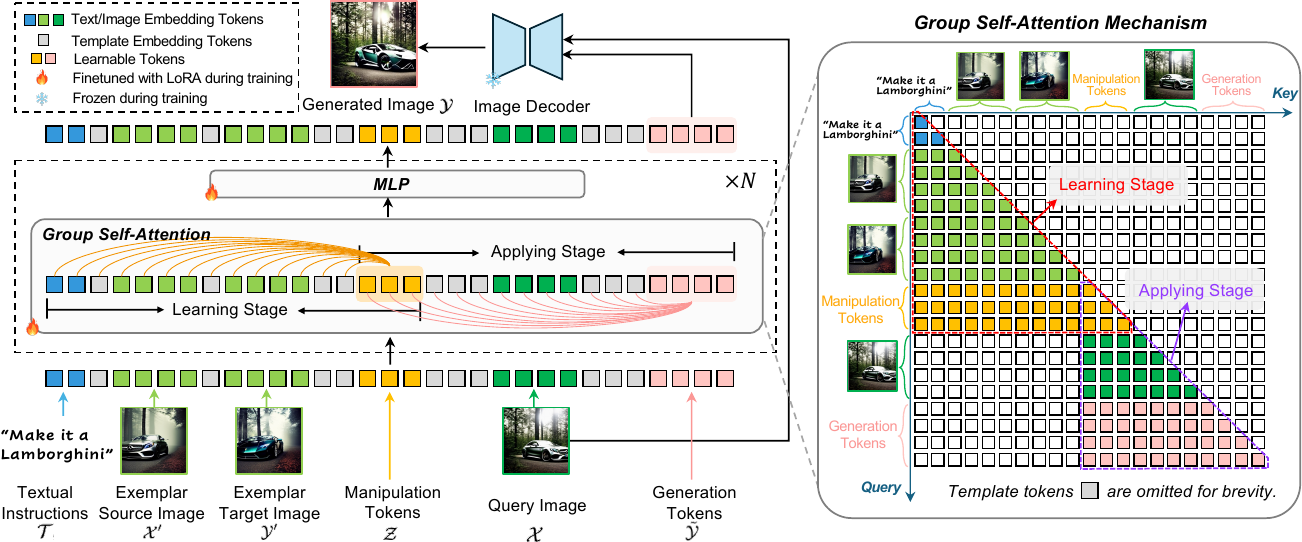}
\vspace{-0.7cm}
\caption{Overview of the proposed InstaManip architecture (left) and group self-attention mechanism (right, represented by query-key matrix). We first tokenize all input texts and images, and fill them in a prompt template with learnable manipulation and generation tokens. We input the prompt into the proposed model which is composed of $N$ blocks. The group self-attention layer in each block learns an explicit manipulation representation $\mathcal{Z}$ and applies it to the new query image. We forward final generation tokens and query image to the image decoder for final image synthesis. In the left part, we only show the self-attention correlations that connect with manipulation tokens or generation tokens for brevity. We also omit encoders, input projection layers and skip connections for simplicity.
}
\vspace{-0.5cm}
\label{fig:model}
\end{figure*}

Previous autoregressive models mix up the two stages when doing in-context learning, thus increasing the problem complexity. In contrast, our model solves this problem in a divide-and-conquer manner. We explicitly split the input prompt into two groups by introducing manipulation tokens (\ie, $\mathcal{Z}$), and then implement self-attention within each group. Our proposed \textit{group self-attention} mechanism (in \cref{sec:gsa}) therefore is able to model learning stage and applying stage separately (following \cref{eq:2stage_formulation}) during end-to-end training, which decomposes the complex in-context learning problem into two easier tasks, thus leading to a better performance (see \cref{fig:gsa_vs_plain}). Furthermore, we introduce a relation regularization strategy (in \cref{sec:realtion_reg}) to guide the model to separate the desired image transformation from irrelevant information for better representation learning.

\subsection{Prompt Composition}

The architecture of our proposed InstaManip model is demonstrated in \cref{fig:model}. Following previous work \cite{ge2024seed,sun2023emu}, we use a pre-trained image encoder to tokenize the exemplar images and the query image each into 64 visual tokens, and then use a linear layer to align visual tokens with the embedding space of autoregressive model. Different from previous methods \cite{bai2024sequential,ge2024seed} that use discrete visual embeddings, we encode images into a continuous space for better representation \cite{li2024autoregressive}. The $M$ learnable manipulation tokens are initialized from pre-trained word embeddings. We construct the full input prompt using the following template. 

\vspace{0.2cm}
\noindent\textbf{Prompt Template:} ``\textit{Here is an image manipulation instruction {\color{NavyBlue} \{textual instruction\}}, which can edit source image {\color{LimeGreen} \{exemplar source image\}} to target image {\color{LimeGreen} \{exemplar target image\}}. The editing is embedded in {\color{Dandelion} \{manipulation tokens\}}. Learn from the instruction with the exemplar pairs and apply the same manipulation to this image {\color{ForestGreen} \{query image\}}.}"

\vspace{0.2cm}
\noindent Note that the manipulation tokens are placed between exemplar images and query image, so that the manipulation features learned with causal mask are independent from the query image. We will elaborate more details in \cref{sec:gsa}. We append 64 learnable generation tokens to the end as the initial state for generating the target manipulated image. Finally, we tokenize all texts in the well-designed prompt with a pre-trained text encoder. The tokenized prompt is then fed into the proposed InstaManip model.

\subsection{Group Self-Attention}
\label{sec:gsa}

As illustrated in \cref{fig:model}, InstaManip is composed of $N$ self-attention blocks. Each block consists of a group self-attention (GSA) layer and a multi-layer perception (MLP). Our key innovation is dividing the input prompt tokens into two groups with the manipulation tokens as the only bridge to connect them. Then we conduct self-attention with causal masks in the two groups separately in a single forward pass.

As demonstrated in \cref{fig:model} (right), in the first group that contains textual instructions $\mathcal{T}$ and exemplar images ($\mathcal{X}',\mathcal{Y}'$), the self-attention is written as
\begin{equation}\label{eq:gsa_stage1}\footnotesize
    GSA_1 = \sigma\left(\frac{\bm{Q}_{\left[\mathcal{T},\mathcal{X}',\mathcal{Y}',\mathcal{Z}\right]} \bm{K}_{\left[\mathcal{T},\mathcal{X}',\mathcal{Y}',\mathcal{Z}\right]}^T - \bm{S}_1}{\sqrt{D}}\right) \cdot \bm{V}_{\left[\mathcal{T},\mathcal{X}',\mathcal{Y}',\mathcal{Z}\right]},
\end{equation}
where we use subscript $\left[\mathcal{T},\mathcal{X}',\mathcal{Y}',\mathcal{Z}\right]$ for $\bm{Q}$, $\bm{K}$, $\bm{V}$ to denote the query, key and value tokens of $(\mathcal{T},\mathcal{X}',\mathcal{Y}',\mathcal{Z})$. To apply causal mask, $\bm{S}_1$ is an upper triangular matrix with values above the diagonal filled with infinity and values at the other locations being zeros. $\sigma$ is the softmax function and $D$ denotes the length of each token. We omit template embedding for brevity in \cref{eq:gsa_stage1}. In this group, manipulation tokens $\mathcal{Z}$ abstract high-level manipulation embeddings from both textual instructions and visual examples, regardless of the query image (\ie, learning stage \circled{1} in \cref{eq:2stage_formulation}). Likewise, in the second group covering the query image $\mathcal{X}$ and the generation tokens $\mathcal{\tilde{Y}}$, the group self-attention is
\begin{equation}\footnotesize
    GSA_2 = \sigma\left(\frac{\bm{Q}_{\left[\mathcal{Z},\mathcal{X},\mathcal{\tilde{Y}}\right]} \bm{K}_{\left[\mathcal{Z},\mathcal{X},\mathcal{\tilde{Y}}\right]}^T - \bm{S}_2}{\sqrt{D}}\right) \cdot \bm{V}_{\left[\mathcal{Z},\mathcal{X},\mathcal{\tilde{Y}}\right]},
\end{equation}
where $\bm{S}_2$ is also an upper triangular matrix akin to $\bm{S}_1$. We constrain the scope of tokens within the second group, so that textual instruction and exemplar images are invisible to generation tokens when they are evolving to the desired output (\ie, applying stage \circled{2} in \cref{eq:2stage_formulation}). The manipulation tokens are the only condition used to manipulate the query image, which hence enforces the model to learn transferrable manipulation features in these tokens. 

The output of GSA is then input into an MLP. Skip connections are applied to GSA and MLP to compensate some missing details. After going through $N$ blocks, the generation tokens and query image are fed into a pre-trained visual decoder to reconstruct the output image after manipulation.

\subsection{Relation Regularization}
\label{sec:realtion_reg}

In group self-attention, the manipulation tokens may still learn misleading features that are unrelated to the desired transformation. To tackle this challenge, we propose a relation regularization strategy to make manipulation embeddings of semantically similar instructions stay close, and keep a proper distance from those of different instructions. Specifically, with a training batch size of $B$, we average the $M$ manipulation tokens $\mathcal{Z}_i\in\mathbb{R}^{B\times M\times D}$ in the $i$-th block to get a single feature vector for each sample, and then apply L2-normalization to each feature vector, resulting in a representation of $\mathcal{\bar{Z}}_i\in\mathbb{R}^{B\times D}$. A relation matrix can be obtained through the inner product of each pair of data samples, \ie, $\mathcal{\bar{Z}}_i\mathcal{\bar{Z}}_i^T\in\mathbb{R}^{B\times B}$, where a greater value implies a closer relation between the two manipulation features. Moreover, the relation of manipulations can be directly represented by the semantic similarity of textual instructions. We utilize a pre-trained CLIP \cite{radford2021learning} text encoder $\phi$ to encode textual instructions and apply L2-normalization to the embedding. Likewise, the relation matrix is then obtained also by inner product, \ie, $\phi(\mathcal{T})\phi(\mathcal{T})^T \in\mathbb{R}^{B\times B}$. We use CLIP encoded relation matrix to regularize the optimization of manipulation tokens by enforcing the two matrices to be close in each GSA layer using MSE loss. The proposed relation regularization strategy is formulated as
\vspace{-0.1cm}
\begin{equation}
    \mathcal{L}_{relation} = \frac{1}{N} \sum_{i=1}^N \left\|\mathcal{\bar{Z}}_i\mathcal{\bar{Z}}_i^T- \phi(\mathcal{T})\phi(\mathcal{T})^T\right\|_F^2,
\vspace{-0.1cm}
\end{equation}
where $\|\cdot\|_F$ is the Frobenius norm which is the square root of the sum of the squares of all matrix elements. Relation regularization encourages our model to learn features directly relevant with the manipulation operation, leading to notable gains in model performance. In addition, we also use MSE as reconstruction loss $\mathcal{L}_{recon}$ between the ground truth image tokens and generation tokens in the final output. The final training loss is a linear combination of reconstruction loss and relation regularization with a coefficient $\alpha$ balancing the two components, which is written as
\begin{equation}
    \mathcal{L} = \mathcal{L}_{recon} + \alpha \mathcal{L}_{relation}.
\end{equation}

\subsection{Implementation Details}

InstaManip exploits the autoregressive architecture of LLaMA-13B \cite{touvron2023llama} consisting of $N=40$ self-attention layers. We use the ViT of Qwen \cite{bai2023qwen} as image encoder and SDXL \cite{podell2024sdxl} as image decoder. We use $M=30$ learnable manipulation tokens in the experiments. In training, we freeze image encoder and decoder, and only optimize group self-attention layers and MLP using LoRA \cite{hu2022lora}. Please refer to Sec. \ref{sec:training_details} in supplementary for more training details.
\vspace{-0.2cm}

\section{Experiments}

\begin{table*}[t]
\footnotesize
\centering
\begin{tabular}{lccccccccc}
\toprule
\multirow{2.5}{*}{Methods}  & \multirow{2.5}{*}{Guidance} & \multicolumn{4}{c}{In Distribution} & \multicolumn{4}{c}{Out of Distribution} \\
\cmidrule(lr){3-6} \cmidrule(lr){7-10}
& & CLIP-Dir & CLIP-Vis & CLIP-T & CLIP-I & CLIP-Dir & CLIP-Vis & CLIP-T & CLIP-I  \\
\midrule
InstructPix2Pix \cite{brooks2023instructpix2pix} & Text Only & 14.47 & 23.42 & 24.58 & \textbf{81.28} & - & - & - & - \\
\hdashline
ImageBrush \cite{sun2023imagebrush}              & Text + Image & 16.42 & 25.03 & 26.45 & 71.98 & 15.70 & 23.89 & 24.34 & 70.68 \\
VISII \cite{nguyen2023visual}                    & Text + Image & 15.85 & 24.91 & 26.10 & 80.10 & 14.69 & 22.95 & 26.14 & 78.10 \\
PromptDiffusion \cite{wang2023context}           & Text + Image & 17.13 & 27.69 & 24.07 & 70.67 & 15.41 & 25.49 & 23.85 & 71.19 \\
\rowcolor[HTML]{FAEBD7} \textbf{InstaManip}      & Text + Image & \textbf{19.81} & \textbf{32.39} & \textbf{27.72} & 80.11 & \textbf{18.27} & \textbf{28.23} & \textbf{26.81} & \textbf{79.71} \\
\bottomrule
\end{tabular}
\vspace{-0.2cm}
\caption{Comparison with prior text-guided image editing model and few-shot image manipulation approaches. InstructPix2Pix only uses textual guidance so that it doesn't belong to either of the two settings. We show the results of InstructPix2Pix under in-distribution setting simply for a direct comparison. The {\color{orange} orange} row refers to our InstaManip model performance.}
\label{tab:sota_cmp}
\vspace{-0.3cm}
\end{table*}

\subsection{Dataset and Metrics}

\textbf{Dataset.} We implement experiments using the dataset collected in the work of InstructPix2Pix \cite{brooks2023instructpix2pix}, which is composed of 313,010 diverse image manipulation instructions. For each instruction, there are 1-4 image pairs (source and target) and corresponding captions. We count the occurrence of each word in the instructions and select 30 keywords with low occurrence as test set candidates. Then we filter out all instructions that contain any of the 30 keywords from the training set, to make sure all test instructions (and their variants) are \textit{invisible} to models during training. 
We further check out the test data and remove the samples with incorrect ground truth. Finally, we end up with 325 instructions and 1296 data samples in the test set. More details are further elaborated in Sec. \ref{sec:test_set} of the supplementary.

\noindent\textbf{Metrics.} We adopt image-to-image similarity, image-to-text similarity, and user study as metrics to measure the model performance. To begin with, we adopt three metrics that are widely used in previous image manipulation studies \cite{sheynin2024emu,brooks2023instructpix2pix,sun2023imagebrush}, including (1) CLIP image-text direction alignment (CLIP-Dir) -- measuring the alignment of image change and caption change, (2) CLIP image-text output similarity (CLIP-T) -- measuring the agreement of manipulated image and output caption, and (3) CLIP image-image similarity (CLIP-I) -- measuring the similarity of query and manipulated images. For the few-shot learning setting, we use (4) visual CLIP similarity (CLIP-Vis) \cite{nguyen2023visual} as an additional metric, which measures the alignment of exemplar image change and query image change. In addition, we conduct (5) user study to collect human preferences on the outputs of our model and all competitors.

\begin{figure*}[t]
\centering
\includegraphics[width=\linewidth]{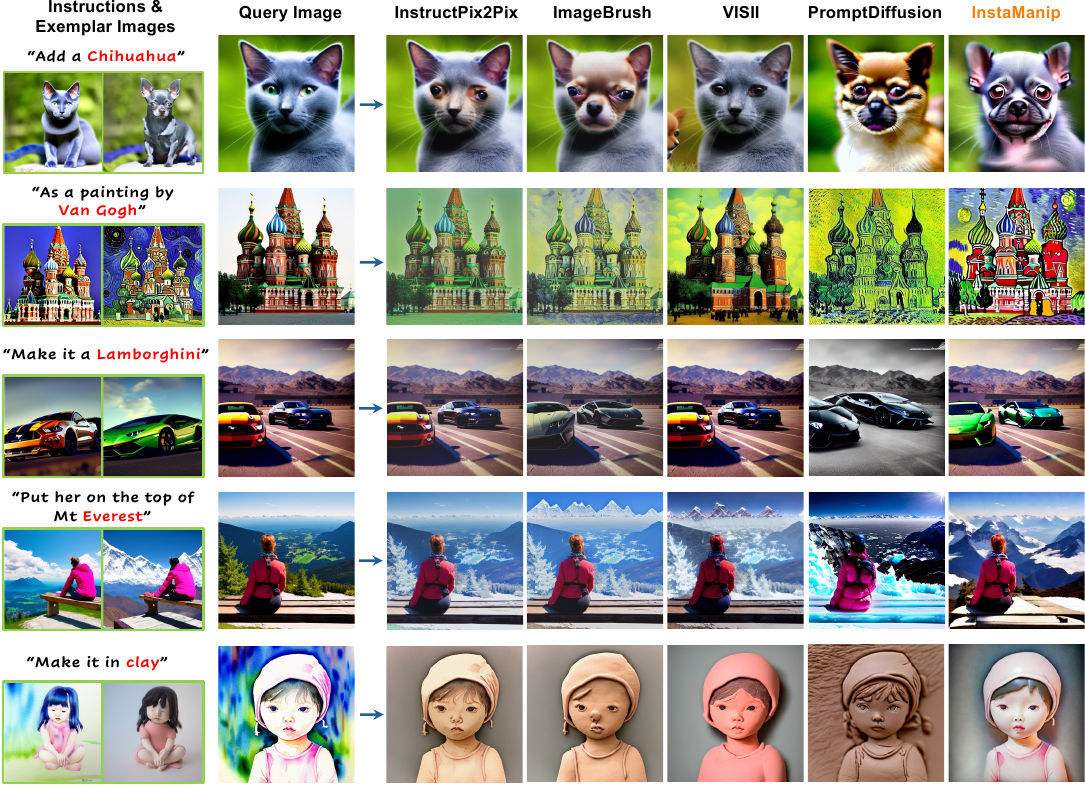}
\vspace{-0.7cm}
\caption{Qualitative comparison with InstructPix2Pix and previous few-shot image manipulation methods. All instructions containing selected keywords (highlighted in {\color{red} red}) are excluded from the training set, so that the models are not optimized on these manipulation operations. Our model follows the textual instruction better, and performs the transformation more aligned with exemplar image pairs.}
\label{fig:cmp_sota}
\vspace{-0.6cm}
\end{figure*}

\subsection{Comparison with Prior Methods}
\label{sec:cmp_sota}

We compare InstaManip with previous few-shot image manipulation models including ImageBrush \cite{sun2023imagebrush}, VISII \cite{nguyen2023visual} and PromptDiffusion \cite{wang2023context}. The three models enable the latent diffusion model to learn from exemplar images by using grid strategy, optimizing the latent condition embedding, and using a separate controller, respectively. We also directly compare with InstructPix2Pix (IP2P) \cite{brooks2023instructpix2pix} trained only using textual instructions. We train the four models on our training set using the default hyperparameters described in their papers. More implementation details are elaborated in Sec. \ref{sec:implementation_previous_methods} of the supplementary. We use two test settings for a thorough comparison -- (1) in-distribution evaluation: exemplar images and query image share the same manipulation instructions and image contents (\eg, scenes, objects, background, \etc); (2) out-of-distribution evaluation: exemplar images and query image share the same transformation, yet different image contents (\eg, indoor scene vs. outdoor scene), which thus makes it more challenging. In this experiment, we use only one exemplar pair for both settings.

Experiment results are demonstrated in \cref{tab:sota_cmp}. In in-distribution setting, all few-shot image manipulation models outperform IP2P in most metrics, suggesting the importance of visual guidance in learning new instructions. In addition, our model further surpasses pervious methods by 2.68\%, 4.70\% and 1.27\% in CLIP-Dir, CLIP-Vis and CLP-T respectively. The prominent improvement in CLIP-Dir and CLIP-Vis indicates that InstaManip follows the textual instructions and visual examples more faithfully, validating the superior in-context learning capability of our model for image manipulation. Though InstaManip lags behind IP2P in CLIP-I, it still achieves the second best performance. We also want to argue that CLIP-I has intrinsic flaws as a metric. A very high CLIP-I score (close to 1) indicates the model does trivial changes to the image, while a low CLIP-I score suggests the model may edit irrelevant areas. This issue makes it hard to assess the performance based on CLIP-I alone. Please refer to Sec. \ref{sec:metric_analysis} in the supplementary for more analysis.

In terms of out-of-distribution setting, it's not surprising to observe an obvious performance drop compared with in-distribution counterpart. However, InstaManip still surpasses the competitors by a great margin in all metrics. The results further suggest that our model learns more transferrable embeddings of desired manipulation, and thus has better generalization capability in more challenging setting.

\begin{figure}[t]
\centering
\includegraphics[width=\linewidth]{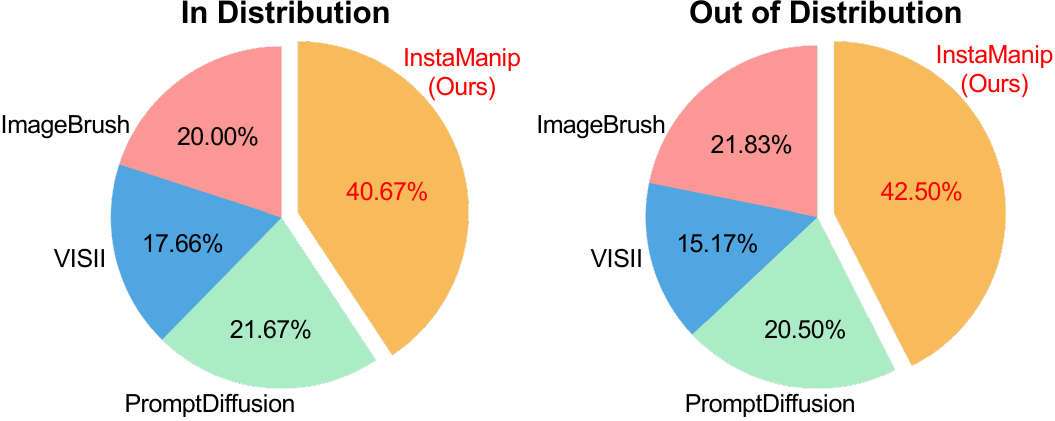}
\vspace{-0.6cm}
\caption{Human evaluation (represented in preference rate) of our model and existing few-shot image manipulation methods.}
\label{fig:user_study}
\vspace{-0.6cm}
\end{figure}

Furthermore, we also conduct user study for a thorough evaluation. As presented in \cref{fig:user_study}, our model surpasses previous methods by a remarkable margin ($\geq$19\%) under the two settings. The results indicate that the output of InstaManip is more aligned with human's subjective criteria, further validating the superiority of our model.

\begin{table}[t]
\small
\centering
\begin{tabular}{cccc}
\toprule
Group SA & Relation Reg. & CLIP-Dir & CLIP-Vis \\
\midrule
\XSolidBrush & \XSolidBrush  & 17.42 & 28.96 \\
\Checkmark   & \XSolidBrush  & 18.96 & 31.08 \\
\rowcolor[HTML]{FAEBD7} \Checkmark & \Checkmark & \textbf{19.81} & \textbf{32.39} \\
\bottomrule
\end{tabular}
\vspace{-0.2cm}
\caption{Evaluation of the contribution of each component. \XSolidBrush \ under Group SA denotes replacing group self-attention layer by vanilla self-attention layer with causal mask. The {\color{orange} orange} row indicates the performance of the full InstaManip model.}
\label{tab:component_ablation}
\vspace{-0.4cm}
\end{table}

\subsection{Visualization for Qualitative Evaluation}

\begin{figure}[t]
\centering
\includegraphics[width=\linewidth]{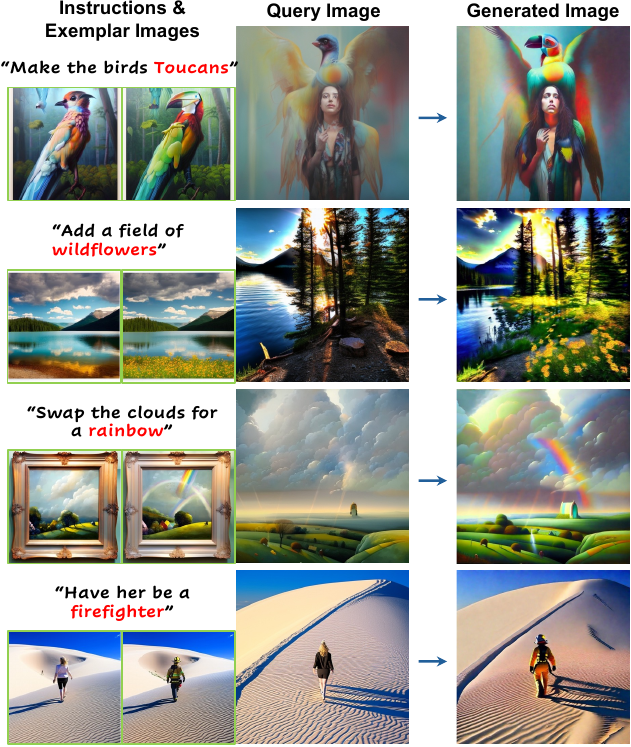}
\vspace{-0.7cm}
\caption{Examples of InstaManip output. Our model learns transformation rules effectively and applies them to new query images.}
\label{fig:demo}
\vspace{-0.6cm}
\end{figure}

The qualitative comparison across previous methods and our model is illustrated in \cref{fig:cmp_sota}. Without visual examples, IP2P may fail to understand instructions that are unseen during training, thus making trivial modification on query images. VISII may also conduct minor transformation (\eg, row1-3), probably due to overfitting in test-time finetuning. ImageBrush and PromptDiffusion understand the instructions better than VISII and make necessary modifications to query images. However, they are still sub-optimal in following visual prompts, so that they may overly edit the images (\eg, row3), or change the images in a distinct direction than visual examples (\eg, row2, row4-5). In contrast, our model implements accurate manipulation aligned with both textual and visual guidance. 
See \cref{fig:demo} for more demonstration.

\begin{table}[t]
\small
\centering
\setlength{\tabcolsep}{0.16cm}
\begin{tabular}{cccc}
\toprule
Textual Instructions & Visual Examples & CLIP-Dir & CLIP-Vis \\
\midrule
\XSolidBrush & \Checkmark  & 16.65 & 28.02 \\
\Checkmark   & \XSolidBrush  & 15.08 & 22.96 \\
\rowcolor[HTML]{FAEBD7} \Checkmark & \Checkmark & \textbf{19.81} & \textbf{32.39} \\
\bottomrule
\end{tabular}
\vspace{-0.2cm}
\caption{Analysis on the impact of textual instructions and visual examples. The {\color{orange} orange} row indicates the result of our full model.}
\label{tab:modality_ablation}
\vspace{-0.4cm}
\end{table}

\subsection{Ablation Study}

\begin{figure}[t]
\centering
\includegraphics[width=\linewidth]{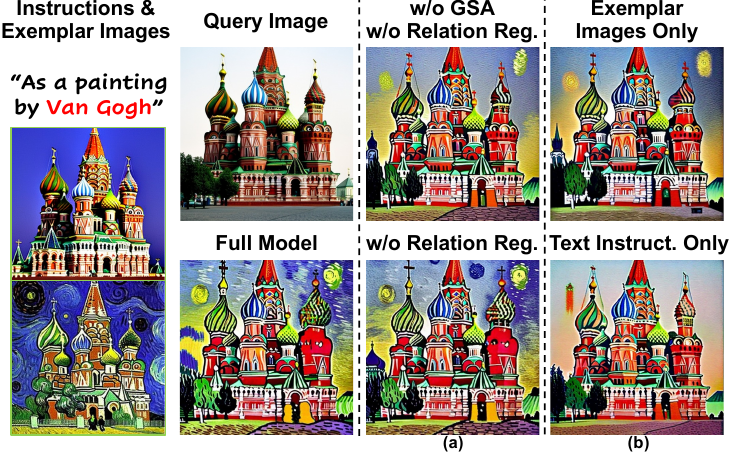}
\vspace{-0.7cm}
\caption{Qualitative evaluation of the contribution of (a) each component, and (b) each modality in the contexts.}
\label{fig:ablation}
\vspace{-0.4cm}
\end{figure}

\textbf{Ablation of Components}. To begin with, we evaluate the contribution of each key component in InstaManip to the final performance. Quantitative results are shown in \cref{tab:component_ablation}. Without using group self-attention and relation regularization, the model is degraded to a plain autoregressive architecture. Using group self-attention alone can improve CLIP-Dir and CLIP-Vis by 1.54\% and 2.12\% respectively. The notable gains validate the effectiveness of modeling learning stage and applying stage separately. After conducting relation regularization in model training, the performance is further boosted by 0.85\% in CLIP-Dir and 1.31\% in CLIP-Vis. This improvement supports our hypothesis that relation regularization can prevent the model from learning irrelevant features by enforcing a structured latent space. Qualitative ablation in \cref{fig:ablation}(a) also presents the progressive improvement of adding the two components.

\vspace{0.1cm}
\noindent\textbf{Ablation of Guidance}. In addition to model components, we also investigate the impact of textual instructions and visual examples on our model. As presented in \cref{tab:modality_ablation} and \cref{fig:ablation}(b), using either textual instructions and visual examples alone results in a significant performance drop. The possible explanation is that textual instructions are more succinct and straightforwad without irrelevant disturbance, while visual examples show more local details that are difficult to describe in texts. They complement each other and thus make the model learn a more robust embedding than using them separately. Another surprising finding is that using visual examples alone leads to a better result than using textual instructions alone. We suspect the reason is that the domain gap between text tokens and image tokens still exists in MLLM feature space. Hence, the image generation tokens can learn from exemplar images more easily than from textual instructions. More analysis and ablation study of our model are shown in Sec. \ref{sec:additional_experiments} of the supplementary.

\begin{figure}[t]
\centering
\includegraphics[width=\linewidth]{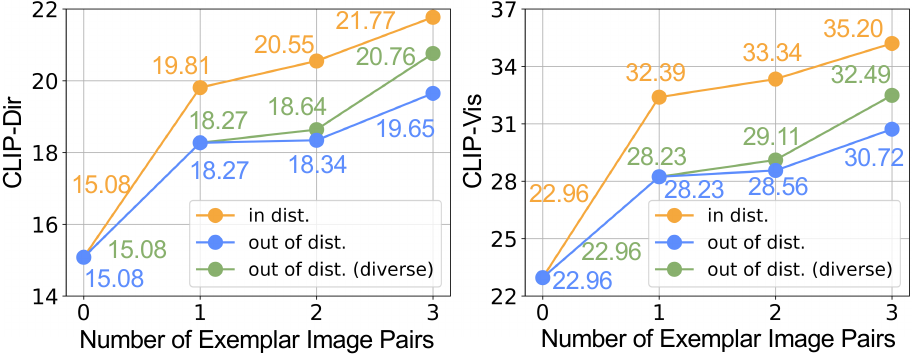}
\vspace{-0.7cm}
\caption{The performance of our model with different numbers of exemplar image pairs. Our model achieves better performance in all the three settings by involving more visual examples.}
\label{fig:scaleup}
\vspace{-0.6cm}
\end{figure}

\subsection{Scaling Up with More Exemplar Images}

We implement extra experiments to study the performance of our model with regard to the number of exemplar image pairs. The results in \cref{fig:scaleup} suggest that the performance of our model is further boosted in both in-distribution and out-of-distribution settings by using more exemplar images. When more than one exemplar image pairs are involved, we introduce a variant of out-of-distribution setting to test the impact of diversity of visual examples. In this setting (\ie, out of dist.(diverse) in \cref{fig:scaleup}, green line), different exemplar images contain distinct scenes, objects and styles, composing a highly diverse visual prompt. In contrast, the contents of exemplar images are very similar in regular out-of-distribution setting (blue line). In \cref{fig:scaleup}, we observe a nontrivial improvement of using diverse visual prompts over the regular setting. It is probably because the high diversity helps the model to better recognize the desired transformation from irrelevant image contents. We also find the gain of increasing examples from 1 to 2 is smaller than from 2 to 3, which is contrary to intuition. Similar phenomenon is also observed in previous work \cite{nguyen2023visual,perez2021true,brown2020language}. The possible reason is that the addition of the third example provides more cues to learn the underlying transformation rules, exactly pushing the model to surpass a representational threshold.

\begin{figure}[t]
\centering
\includegraphics[width=\linewidth]{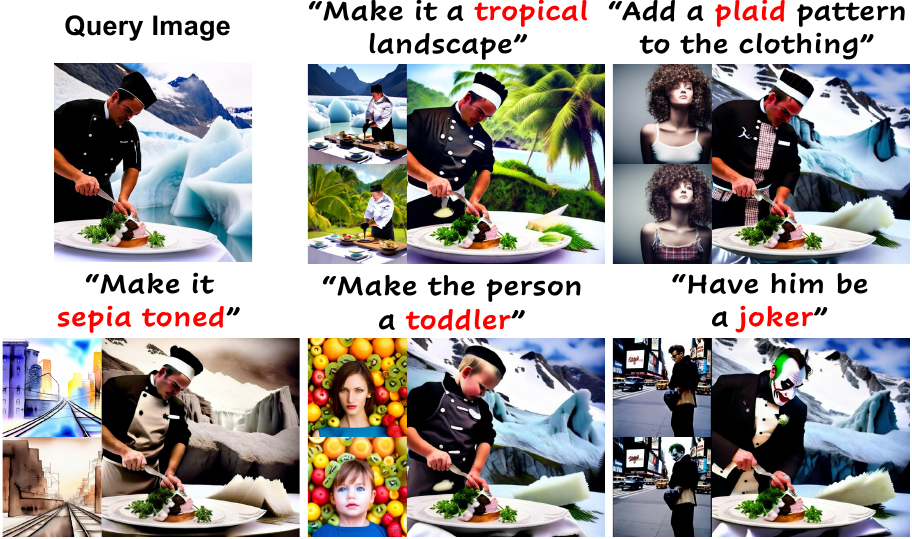}
\vspace{-0.6cm}
\caption{Demonstration of different manipulation on the same query image. Our model successfully edits the image conditioned on various textual and visual guidance.}
\label{fig:gen_same_query}
\vspace{-0.6cm}
\end{figure}

\subsection{Various Manipulation on the Same Image}

Besides pre-defined instructions in the dataset, we further validate the generalization capability of our model to new image-instruction pairs. As illustrated in \cref{fig:gen_same_query}, we ask the model to edit the same query image using various textual instructions coupled with exemplar images. Our model effectively learns the desired transformation rule from the examples, and correctly applies it to the query image.

\section{Conclusion}
In this paper, we propose InstaManip, an autoregressive model consisting of novel group self-attention layers for few-shot image manipulation. Inspired by human's learning process, the key intuition of our approach is breaking down the in-context learning paradigm into learning and applying stages, and modeling the two stages separately in the end-to-end training. We also adopt a relation regularization strategy to identify the underlying manipulation rules from undesired visual features. Detailed experiments demonstrate a notable improvement of InstaManip over prior methods, as well as the scalability of our model. Our work is an important attempt to solve few-shot image manipulation problem with novel design in autoregressive architectures, which paves the way for improving generic in-context learning capability of autoregressive models in various visual tasks.

{
    \small
    \bibliographystyle{ieeenat_fullname}
    \bibliography{main}
}

\maketitlesupplementary
\appendix

This is the supplementary material for the submission titled ``Unleashing In-context Learning of Autoregressive Models for Few-shot Image Manipulation''. We organize the content as follows:

\noindent\textbf{\hyperref[sec:metric_analysis]{A} -- Defects of CLIP-I as a Metric} \\
\textbf{\hyperref[sec:additional_experiments]{B} -- Additional Experiment Results} \\
\hyperref[sec:ablation_token_number]{B.1} -- Analysis on the Number of Manipulation Tokens \\
\hyperref[sec:same_txt_diff_vis]{B.2} -- Manipulation with the Same Textual Instruction and Different Exemplar Images  \\
\hyperref[sec:cmp_generic_autoregressive]{B.3} -- Comparison with the Generic Autoregressive Model  \\
\hyperref[sec:additional_visualization]{B.4} -- Additional Visualization  \\
\hyperref[sec:failures]{B.5} -- Failure Cases  \\
\textbf{\hyperref[sec:more_implementation]{C} -- Implementation Details} \\
\hyperref[sec:test_set]{C.1} -- Establishment of Test Set \\
\hyperref[sec:training_details]{C.2} -- Training Details of Our Model \\
\hyperref[sec:implementation_previous_methods]{C.3} -- Implementation of Previous Methods \\
\hyperref[sec:userstudy_details]{C.4} -- Details of User Study \\
\textbf{\hyperref[sec:limitation_future_work]{D} -- Limitation and Future Work} \\
\textbf{\hyperref[sec:release]{E} -- Code and Data Release}

\renewcommand\thesection{\Alph {section}}
\renewcommand\thesubsection{\thesection.\arabic{subsection}}

\section{Defects of CLIP-I as a Metric}
\label{sec:metric_analysis}

\begin{figure}[t]
\centering
\includegraphics[width=\linewidth]{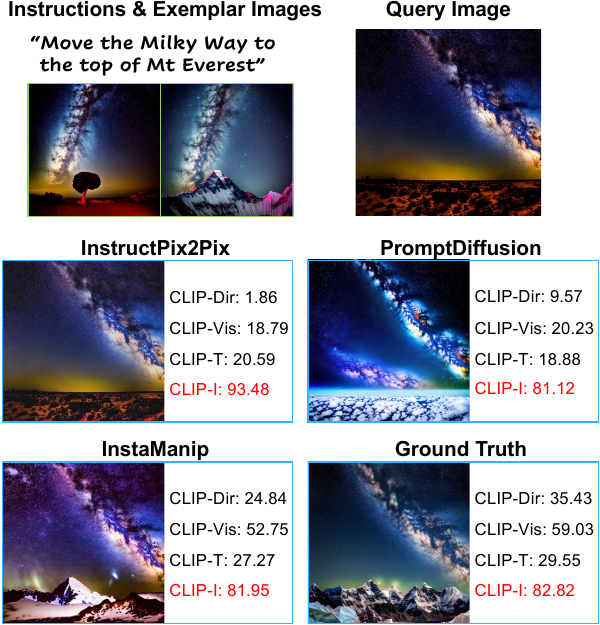}
\caption{Comparison of the four CLIP-based metrics on the outputs of three models and the ground truth. CLIP-I is highlighted in {\color{red} red}. Please refer to Sec. \ref{sec:metric_analysis} for the explanation.}
\label{fig:metric_analysis}
\end{figure}

In \cref{sec:cmp_sota} of the main paper, we argue that CLIP-I (similarity between the query image and the manipulated image) has inherent defects when used as a metric for image manipulation. In order to further explain the reason, we calculate the four CLIP-based metrics used in our experiments (CLIP-Dir, CLIP-Vis, CLIP-T, CLIP-I) on the outputs of three models and the ground truth, which is shown in \cref{fig:metric_analysis}. 

Compared with InstructPix2Pix \cite{brooks2023instructpix2pix} and PromptDiffusion \cite{wang2023context}, our model follows the textual and visual guidance more faithfully in this instance. Nevertheless, this advantage is not correctly reflected by the CLIP-I metric. InstructPix2Pix conducts a trivial modification to the query image, thus resulting in a high similarity between the query image and the output. It's worth noting that the CLIP-I score of InstructPix2Pix is even higher than the score of the ground truth. In contrast, PromptDiffusion overly edits the query image, leading to a CLIP-I score lower than InstaManip and ground truth. Our model (which has the best performance) and ground truth have medium CLIP-I scores between InstructPix2Pix and PromptDiffusion. This example suggests that a higher or lower CLIP-I score does not necessarily correspond to a better performance in the image manipulation task. Hence, it's hard to accurately compare the performance of two methods based on CLIP-I alone. Fortunately, the other three metrics correctly discriminate the performance of the three models, so we use them as the primary metrics in our experiments.

\section{Additional Experiment Results}
\label{sec:additional_experiments}

\begin{table}[t]
\small
\centering
\setlength{\tabcolsep}{0.25cm}
\begin{tabular}{ccc}
\toprule
\# Manipulation Tokens & CLIP-Dir & CLIP-Vis \\
\midrule
10 & 18.24 & 29.87 \\
20 & 19.07 & 31.10 \\
\rowcolor[HTML]{FAEBD7} \textbf{30} & \textbf{19.81} & \textbf{32.39} \\
40 & 19.74 & 32.21 \\
50 & 19.66 & 32.20 \\
\bottomrule
\end{tabular}
\caption{Analysis on the impact of the number of manipulation tokens. The {\color{orange} orange} row indicates our final model. Pleae refer to Sec. \ref{sec:ablation_token_number} for the explanation.}
\label{tab:token_number_ablation}
\end{table}

\begin{figure*}[t]
\centering
\includegraphics[width=\linewidth]{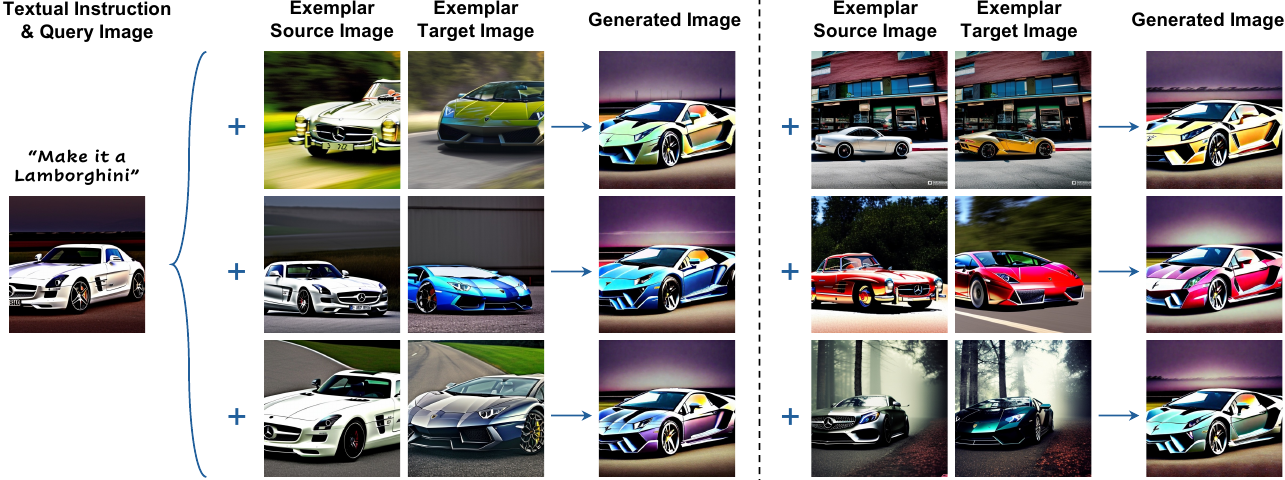}
\caption{The visualization of manipulating the query image using the same textual instruction, yet different visual examples. When we use exemplar target images of Lamborghini with different colors, our model successfully captures this local feature from the visual guidance, and changes the colors in the generated images accordingly. Please refer to Sec. \ref{sec:same_txt_diff_vis} for the detailed analysis.}
\label{fig:same_txt_diff_vis}
\end{figure*}

\subsection{Analysis on the Number of Manipulation Tokens}
\label{sec:ablation_token_number}

We implement experiments to validate the impact of different numbers of manipulation tokens. \cref{tab:token_number_ablation} shows that the performance is boosted by increasing the number of manipulation tokens from 10 to 30. If more than 30 tokens are used in our model, the performance remains comparable to that observed with 30 tokens, suggesting that the model has reached a saturation point. Consequently, we set the number of manipulation tokens as 30 in the final InstaManip model.

\subsection{Manipulation with the Same Textual Instruction and Different Exemplar Images}
\label{sec:same_txt_diff_vis}

One benefit of using exemplar images in image manipulation is that the images effectively convey the desired local details to the model, which may be missing in textual instructions. To validate if the proposed model can effectively learn the visual features, we apply our model to a given image using the same textual instruction yet different visual examples. The results are illustrated in \cref{fig:same_txt_diff_vis}. In this experiment, we use different exemplar pairs following the same textual instruction. The major difference of these examples is the color of the Lamborghini in the exemplar target images. Our model learns this visual feature and successfully edits the query image using similar colors, which exactly reflects the advantage of few-shot image manipulation.

\begin{table}[t]
\footnotesize
\centering
\setlength{\tabcolsep}{0.15cm}
\begin{tabular}{lccccc}
\toprule
Methods  & Guidance  & CLIP-Dir & CLIP-Vis & CLIP-T & CLIP-I \\
\midrule
\multicolumn{6}{c}{\textit{In Distribution}}\\
\midrule
Emu2 \cite{sun2024generative}                 & Text + Image & 15.26  & 24.64  & 27.02  & 76.89 \\ 
\rowcolor[HTML]{FAEBD7} \textbf{InstaManip}   & Text + Image & \textbf{19.81} & \textbf{32.39} & \textbf{27.72} & \textbf{80.11} \\
\midrule
\multicolumn{6}{c}{\textit{Out of Distribution}}\\
\midrule
Emu2 \cite{sun2024generative}                 & Text + Image & 14.09  & 21.65  & 20.17  & 65.80 \\
\rowcolor[HTML]{FAEBD7} \textbf{InstaManip}   & Text + Image & \textbf{18.27} & \textbf{28.23} & \textbf{26.81} & \textbf{79.71} \\
\bottomrule
\end{tabular}
\caption{Comparison with Emu2. InstaManip outperforms the generic autoregressive model by a great margin. Additional discussions are shown in Sec. \ref{sec:cmp_generic_autoregressive}.}
\label{tab:cmp_generic_autoregressive}
\end{table}

\subsection{Comparison with the Generic Autoregressive Model}
\label{sec:cmp_generic_autoregressive}

In this paper, we propose an autoregressive model with enhanced in-context learning capability for few-shot image manipulation. Prior to our work, there is some work about using the autoregressive architecture as a generic in-context learner for various tasks. Emu2 \cite{sun2024generative} is one of the recent studies in this field, showing awesome performance in visual understanding and image generation problems. We compare our model with Emu2 on few-shot image manipulation. The results are reported in \cref{tab:cmp_generic_autoregressive}. InstaManip greatly surpasses Emu2 across all metrics in both evaluation settings. Despite the existence of generic in-context learners, the result suggests that few-shot image manipulation is still a challenging problem that requires specific novel model design. It also validates the necessity of investigating how to improve in-context learning performance for specific tasks like our work.

\begin{figure*}[t]
\centering
\includegraphics[width=\linewidth]{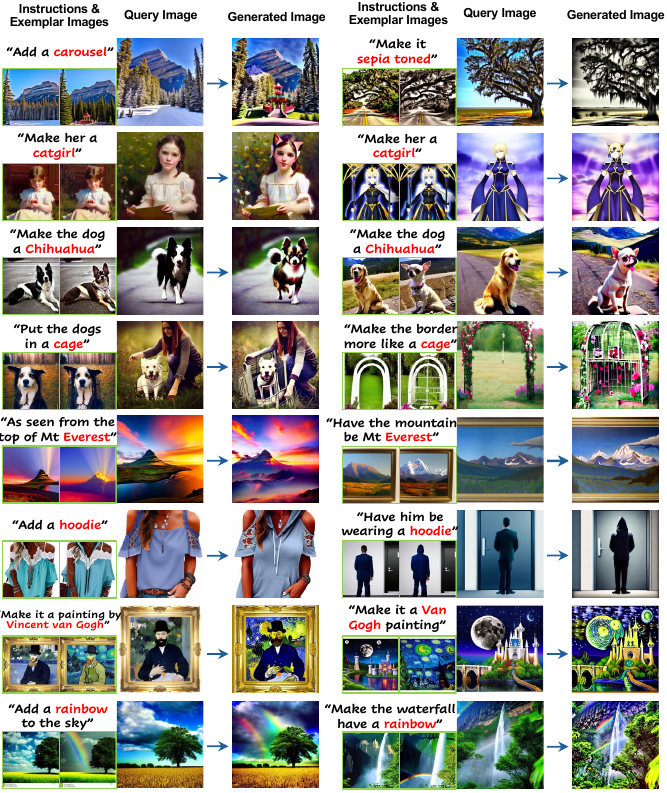}
\caption{Additional visualization of the output from InstaManip. All instructions containing selected keywords (highlighted in {\color{red} red}) are excluded from the training set. Our model learns unseen image manipulation operations from both textual and visual guidance, and applies the learned transformations to the new query images. More examples are presented in \cref{fig:even_more_demo}. See Sec. \ref{sec:additional_visualization} for the discussions.}
\label{fig:more_demo}
\end{figure*}

\begin{figure*}[t]
\centering
\includegraphics[width=\linewidth]{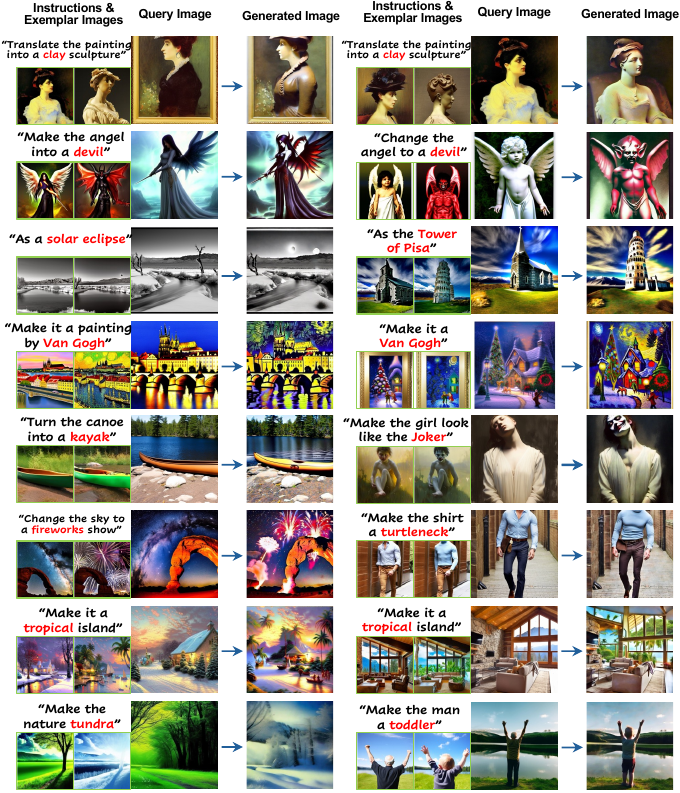}
\caption{More demonstration of the output from InstaManip (continuation of \cref{fig:more_demo}). All instructions containing selected keywords (highlighted in {\color{red} red}) are removed from the training set. Our model edits the query image aligned with both textual instructions and exemplar images. See Sec. \ref{sec:additional_visualization} for the discussions.}
\label{fig:even_more_demo}
\end{figure*}

\subsection{Additional Visualization}
\label{sec:additional_visualization}

To further demonstrate the performance of the proposed InstaManip, we illustrate more outputs from our model in \cref{fig:more_demo,fig:even_more_demo}. By learning an explicit manipulation embedding, InstaManip successfully captures the underlying image transformations from textual and visual guidance, and implements them to the query images faithfully.

\begin{figure*}[t]
\centering
\includegraphics[width=0.9\linewidth]{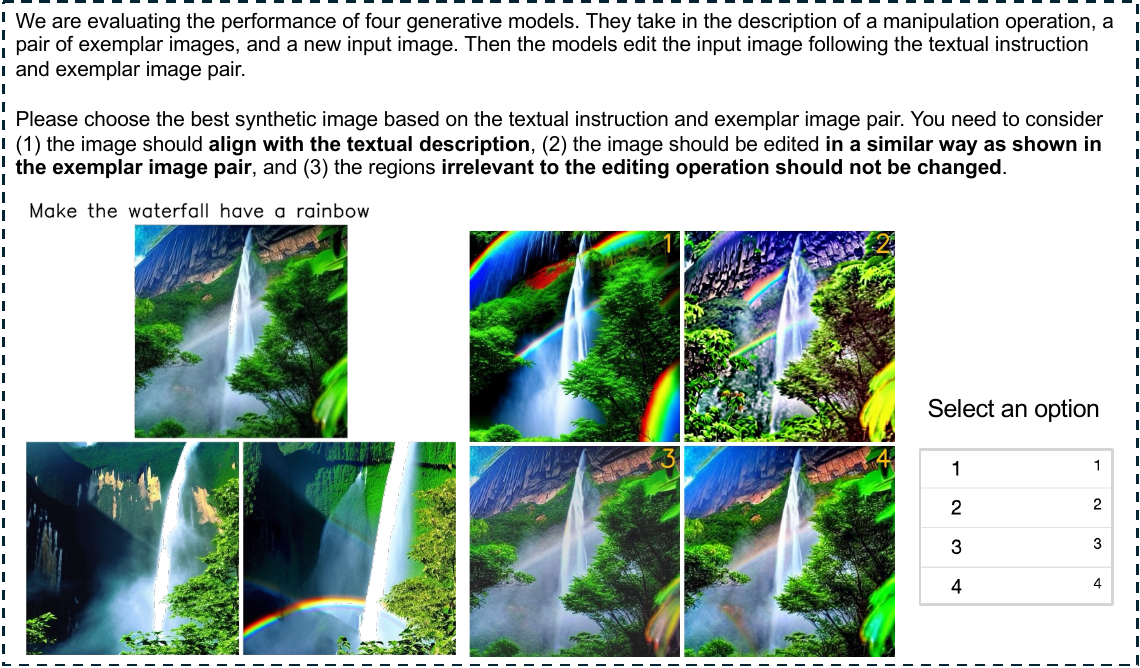}
\caption{The interface used for human evaluation. The four manipulated images are randomly shuffled to avoid potential bias. Please refer to Sec. \ref{sec:userstudy_details} for the detailed elaboration.}
\label{fig:userstudy_interface}
\end{figure*}

\begin{figure}[t]
\centering
\includegraphics[width=\linewidth]{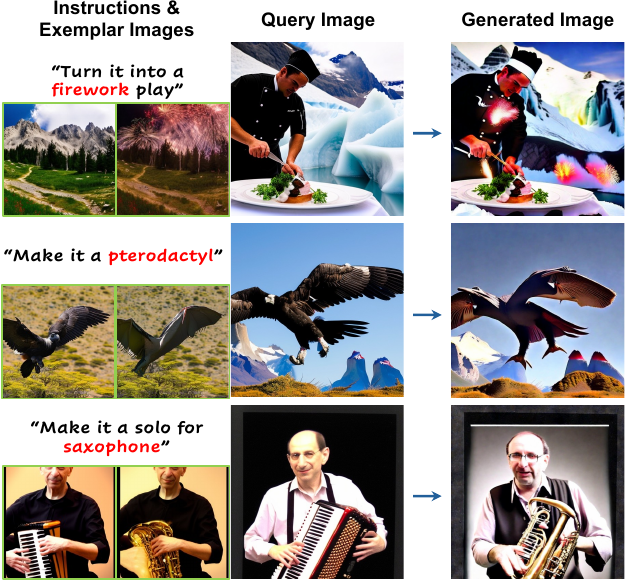}
\caption{Failure cases of InstaManip. Please refer to Sec. \ref{sec:failures} for the discussions.}
\label{fig:failures}
\vspace{-0.3cm}
\end{figure}

\subsection{Failure Cases}
\label{sec:failures}

Though InstaManip shows strong in-context learning capability in image manipulation, we still find it may fail in some cases, as presented in \cref{fig:failures}. To begin with, our model still struggles with the big domain gap between the exemplar images and the query image. In the first example of \cref{fig:failures}, the exemplar images show a view of mountains with plants, while the query image is a picture of a cook preparing meals. Our model places the fireworks in an incorrect position in the generated image. In addition, our model is very likely to fail if the exemplar images do not show the desired visual features accurately. In the second example, the exemplar target image does not show the shape, structure and texture of pterodactyl clearly, thus misleading our model into making a random transformation to the query image. In the third example, the saxophone has a complex structure and texture. Our model fails to accurately capture these subtle details in the generated image. These weaknesses can motivate future investigations into novel models with stronger in-context learning capability. Please refer to Sec. \ref{sec:limitation_future_work} for more discussions.

\section{Implementation Details}
\label{sec:more_implementation}

\subsection{Establishment of Test Set}
\label{sec:test_set}

In order to test our model on unseen instructions, we establish the test set based on selected keywords. Specifically, we count the occurrence of each word in the InstructPix2Pix dataset \cite{brooks2023instructpix2pix}, and select 30 keywrods with low occurrence. The 30 keywords include boxing, cage, carousel, catgirl, Chihuahua, clay, devil, Everest, firefighter, firework, hoodie, joker, kayak, Lamborghini, Lego, Monet, plaid, pterodactyl, rainbow, saxophone, sepia toned, solar eclipse, toddler, toucan, tower of pisa, tropical, tundra, turtleneck, Van Gogh and wildflower. We check out each instance of these keywords manually to filter out low-quality data and incorrect ground truth. The remaining data is used as the test set. We also exclude all instructions that contain any of these selected keywords from the training data, to make sure none of the models is optimized on these keywords in the experiments.

\subsection{Training Details of Our Model}
\label{sec:training_details}

We interpolate the images to a resolution of $448\times448$ before forwarding them to the image encoder. The coefficient $\alpha$ in the loss is set as 0.1. We train our model using the AdamW optimizer \cite{loshchilov2017decoupled} for 20000 iterations on 8 GPUs of NVIDIA A100-SXM4-80GB for 6 days. The batch size is set as 480. We warm up the model to a learning rate of $10^{-4}$ in the first 500 iterations, and reduce the learning rate by cosine annealing in the remaining steps. The weight decay, $\beta_1$ and $\beta_2$ of AdamW are set as 0.05, 0.9 and 0.98 respectively.

\subsection{Implementation of Previous Methods}
\label{sec:implementation_previous_methods}

InstaManip is compared with four models in the main paper \cref{sec:cmp_sota}: InstructPix2Pix \cite{brooks2023instructpix2pix}, ImageBrush \cite{sun2023imagebrush}, VISII \cite{nguyen2023visual} and PromptDiffusion \cite{wang2023context}. As a baseline of text-guided image editing model, InstructPix2Pix is trained only with textual instructions. The model weights are also used for VISII, which relies on a pre-trained InstructPix2Pix model for test-time finetuning. We freeze the weights of InstructPix2Pix and finetune a learnable instruction embedding for each test instance as described in the VISII paper. In contrast, ImageBrush and PromptDiffusion can be trained in an end-to-end way. We train the two models on our training set following the default hyperparameters specified in their work. For a fair comparison, we use both textual instructions and visual examples for VISII, ImageBrush and PromptDiffusion.

\subsection{Details of User Study}
\label{sec:userstudy_details}

We implement human evaluation across our model and the three prior few-shot image manipulation models in the main paper \cref{sec:cmp_sota}. We sample 100 examples from the test set for evaluation. For each sample, we show the textual instruction, exemplar images, query image and the outputs from the four models to human raters. The raters are asked to select the best output image based on three criteria: (1) alignment with the textual instruction, (2) alignment with the exemplar image pair and (3) preservation of irrelevant regions. Each instance is evaluated by six raters. The human evaluation is conducted on Amazon Mechanical Turk. The interface is illustrated in \cref{fig:userstudy_interface}.

\section{Limitation and Future Work}
\label{sec:limitation_future_work}

In this paper, we propose a novel autoregressive architecture to model the learning stage and applying stage separately in in-context learning. Despite the superiority over existing approaches, we still find there are some problems that are not solved by our model. Our model suffers from an obvious performance drop when there is a big gap between the query image and exemplar images. Learning a new object with complex textures is also challenging. Our model may fail to fully capture the subtle details in the visual examples. The failure cases and analysis are elaborated in Sec. \ref{sec:failures}.

In addition to the limitation, our work also points out several valuable research directions.

\begin{itemize}
    \item Addressing cases with significant gap between the query image and visual examples is crucial for real-world applications. Innovative approach for this problem and large datasets containing such out-of-distribution examples are required in future studies.
    \item The dataset used in our work provides four instances at most for each instruction, which prevents us from exploring the saturation point of out model capability by using more than three exemplar pairs in the experiments. More efforts are demanded to build a dataset specifically for few-shot image manipulation.
    \item While our model has shown strong in-context learning capability on image manipulation problem, how to exploit our method for other problems remains to be explored. We expect more future investigations of our findings for stronger generic in-context learning across various tasks.
\end{itemize}

\section{Code and Data Release}
\label{sec:release}

We will release our code, model weights and test set online to the research community to facilitate future studies.

\end{document}